\documentclass{article}

\PassOptionsToPackage{numbers,square,sort&compress}{natbib}
\usepackage[dblblindworkshop,final]{neurips_2026}
\workshoptitle{Interpretability as a Science}
\usepackage[utf8]{inputenc}
\usepackage[T1]{fontenc}
\usepackage{amsfonts}
\usepackage{amsmath}
\usepackage{mathtools}
\usepackage{amssymb}
\usepackage{graphicx}
\usepackage{booktabs}
\usepackage{algorithm}
\usepackage{algpseudocode}
\usepackage{placeins}
\usepackage{microtype}
\usepackage{url}
\usepackage{hyperref}
\usepackage[noabbrev,capitalize]{cleveref}
\usepackage{csquotes}
\usepackage[mathcal]{euscript}
\hypersetup{hidelinks}

\DeclareMathOperator*{\argmax}{arg\,max}
\DeclareMathOperator{\Rot}{Rot}

\title{Halting in Recurrent Sequence Models}
\author{Simon Dr{\"a}ger \\ Salk Institute for Biological Studies}

\begin{document}
\maketitle

\begin{abstract}
Recurrent sequence models, such as recurrent neural networks (RNNs) and state space models (SSMs), are often trained to emit an end-of-sequence (EOS) token when their output sequence is complete. During training, EOS is treated as yet another token without a special loss term or incentive. Although this training paradigm is widely used, it is not understood how models decide to output EOS. We study a minimal version of the EOS halting problem in which networks are trained to emit a given input sequence an instructed number of times, followed by EOS. We train tanh RNNs and linear SSMs on this task, finding that the RNNs encode the number of tokens left to copy in a linear subspace, of which two principal components capture most of the variation. Furthermore, we find that hidden states with the same remaining count converge to a shared representation regardless of their instructed repeat count. By modulating the hidden state using an affine transformation at the boundary between copied sequences, we shift the RNN's state to the representation shared by states with one fewer copy remaining. The unchanged network emits EOS one copy earlier while reproducing the shortened target sequence in \(94.4\%\) of new sequences, revealing that this representation causally determines the stopping time. In the SSMs, the difference between trajectories with different repeat counts is separated across two-dimensional rotation planes whose contributions to the EOS comparison cancel during copying and align near the stopping time. Rotating the planes independently removes most of the intended stopping while preserving the copied content, which shows that the stopping time depends on the planes' relative phases. These results identify two candidate primitives of EOS halting in recurrent models: a shared encoding of the number of remaining tokens in tanh RNNs and phase alignment across fixed rotations in linear SSMs.
\end{abstract}

\section{Introduction}
\label{sec:introduction}

How does a recurrent neural network (RNN) know when to stop outputting new tokens when trained on a task with a variable sequence length? In such a task setting, RNNs commonly end generation by emitting an end-of-sequence (EOS) token after their content. Given an input \(x_t\) and hidden state \(h_{t - 1}\), the model updates its state and assigns a score to every output token according to
\begin{equation*}
  h_t = F_\theta(h_{t - 1}, x_t),
  \qquad
  z_t = W_y h_t.
\end{equation*}
Here \(F_\theta\) is the recurrent update with parameters \(\theta\), \(W_y\) is the output matrix (decoder), and \(z_t\) is a vector containing the scores for each token \(v\) in the task's vocabulary \(\mathcal{V}\). The EOS margin
\begin{equation}\label{eq:eos-margin}
  m_t = z_{t,\mathrm{EOS}} - \max_{v \ne \mathrm{EOS}} z_{t,v}
\end{equation}
provides us with a mathematical object that defines when EOS is output: the model emits content while \(m_t \le 0\) and emits EOS once \(m_t > 0\). When subsequent inputs are fixed, the current state determines the entire future trajectory and can therefore determine the stopping time even while its current margin remains negative. The margin only records when EOS is emitted. Why its sign changes at a particular step has to be explained by the hidden dynamics, which carry the count instruction until then. Existing attempts at characterizing EOS prediction do not provide a transition-level mechanism of how a count instruction is transported into the EOS comparisons (\cref{sec:related-work}).

The \emph{repeat-copy} task augments the copy task by adding a count instruction signaling that the sequence to be copied should be copied \(K \in \mathbb{N}\) times, followed by EOS~\citep{graves2014ntm}. Under different count instructions \(J\) and \(K\), \(J \ne K\), the same model produces hidden trajectories \(h_t^J\) and \(h_t^K\). At each output step shared by such trajectories, the \emph{trajectory difference}
\begin{equation*}
  \Delta_t^{J K} = h_t^J - h_t^K
\end{equation*}
isolates the effect of the repeat count, since both trajectories store the same sequence. Its evolution describes how the count reaches the EOS comparisons.

\begin{figure}[t]
  \centering
  \includegraphics[width=\textwidth]{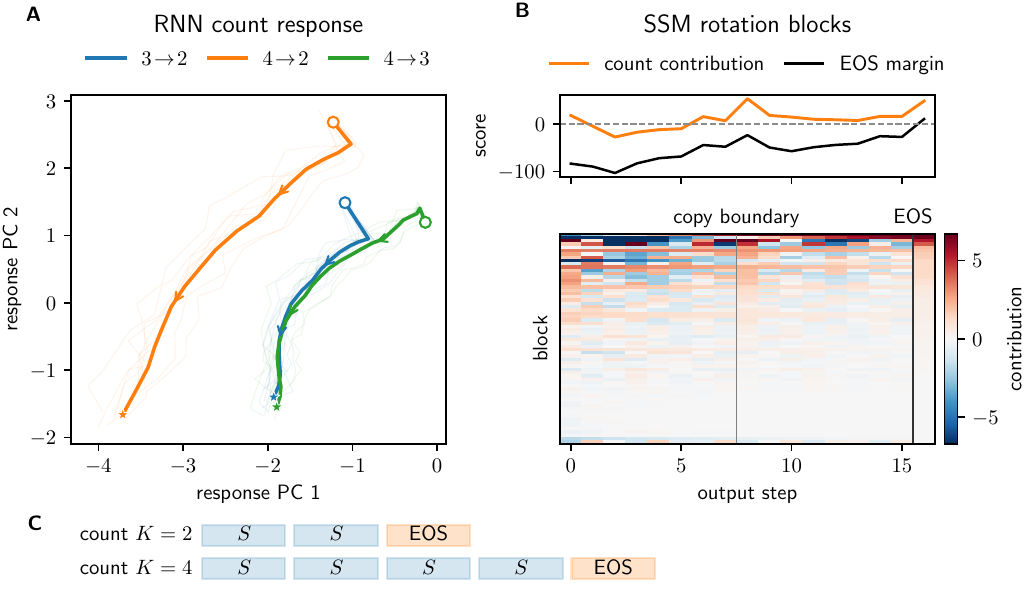}
  \caption{The count instruction changes the EOS time while the stored sequence
  remains fixed. \textbf{A}, RNN trajectory differences on held-out sequences projected onto
  two principal components obtained from separate sequences. Open circles mark
  the state after the count instruction, stars mark the earlier target EOS, and
  arrows follow the autonomous recurrence. \textbf{B}, mean contributions of the
  64 SSM rotation blocks to the limiting EOS comparison for count change
  \(4 \to 2\). Between copy boundaries the contributions have mixed signs. At
  the target EOS, most contributions have the same sign and the EOS margin
  crosses zero. \textbf{C}, repeat-copy examples ending after two or four copies
  of \(S\). RNN seed 5, the model with the median energy in the two plotted components, and SSM seed 1 are shown.}
  \label{fig:opening-mechanisms}
\end{figure}

We consider a minimal RNN architecture and show that its states carry the number of output steps remaining, which at the fixed sequence length is equivalent to the position within a copy together with the number of copies remaining. Trajectories with the same number of remaining steps approach the same states regardless of the count instruction. Within each copy, \(W_h\) mixes \(\Delta_t^{J K}\), and the tanh activation then attenuates each of its components according to the current pair of hidden states. An affine state transformation learned from separate sequences shifts a state at a copy boundary to the states with one fewer copy remaining, which causes EOS to be emitted one copy earlier without changing the tokens output until that point. For comparison, experiments using a linear SSM reveal that \(\Delta_t^{J K}\) evolves within two-dimensional rotation planes whose contributions to the EOS comparisons cancel during copying and increase near the correct stopping time. Independently rotating these contributions removes most of the intended stopping behavior while preserving the copied content (\cref{fig:opening-mechanisms}).

\section{Related work}%
\label{sec:related-work}

Prior work on halting using an EOS token has examined sequence probability, decoding criteria, learned stopping, and length generalization~\citep{sountsov2016length,murray2018correcting,yang2018breaking,ma2019learning}. \citet{newman2020eos} find that EOS training organizes recurrent states into length manifolds and terminal attractors. \citet{grant2025emergent} show that in models trained on tasks using a latent count variable, hidden state interventions on the network's representation of the count variable leads to altered output behavior.

Adaptive computation methods add an explicit halting variable that controls the
number of internal updates a model makes along its trajectory~\citep{graves2016act,figurnov2018probabilistic,banino2021pondernet,dehghani2019universal,elbayad2020depth}.
Structured state space models (SSMs) implement recurrent sequence processing through constrained linear transitions~\citep{gu2022s4,smith2023s5,orvieto2023resurrecting,gu2023mamba}, including the scaled rotation blocks in the linear SSM studied here.

Controlled sequence tasks reveal memory, position, and algorithmic structure in recurrent states~\citep{elman1990finding,karpathy2015visualizing,grefenstette2015learning,joulin2015inferring,henaff2016recurrent,karuvally2023episodic,csordas2024recurrent}. Fixed points and local linearizations describe their evolution under the recurrent transition~\citep{sussillo2013opening,haviv2019understanding,maheswaranathan2019reverse,smith2021jacobian}. Causal interventions test the behavioral role of variables identified by linear decoders~\citep{hewitt2019structural,hewitt2019control,elazar2021amnesic,geiger2021causal}.

\section{Stopping as a hitting-time problem}%
\label{sec:setup}

\paragraph{Task.} We use the repeat-copy task~\citep{graves2014ntm}, in which each input sequence consists of \(L = 8\) symbols \(S = (s_0, \ldots, s_{L - 1})\) with \(s_i \in \{-1, +1\}^4\), yielding sixteen possible content tokens. After the symbols and a delimiter, an instruction \(c_K \in \{-1, +1\}^3\) with a single \(+1\) entry encodes \(K \in \{2, 3, 4\}\) and requests the output
\begin{equation*}
  \underbrace{S\,S\,\cdots\,S}_{K\ \mathrm{copies}}\,\mathrm{EOS}.
\end{equation*}
Unlike \citet{graves2014ntm}, who supply the repeat count as a scalar on a separate input channel, we present it as a one-hot token after the delimiter.

\paragraph{Models.} We use two models: an RNN with one hidden layer and tanh activation, and a linear SSM. Both models have width \(d = 128\). A matrix \(W_x\) first maps each input into the hidden state before the bias-free output matrix \(W_y\) returns the token scores. The recurrent operator is \(W_h\) for the RNN and \(A\) for the SSM:
\begin{align}
  h_t   &= \tanh(W_h h_{t - 1} + W_x x_t), & z_t &= W_y h_t, \label{eq:rnn} \\
  h_t   &= A h_{t - 1} + W_x x_t,          & z_t &= W_y h_t, \label{eq:ssm}
\end{align}
The SSM transition contains 64 learned and scaled \(2 \times 2\) rotation blocks. We train five seeds for each architecture, using \(100{,}000\) updates for the RNNs and \(5{,}000\) updates for the SSMs. Every checkpoint emits EOS at the intended step on the held-out sequences used in the analyses below, and only RNN seed 4 miscopies a single content token, in one of 512 sequences.

\paragraph{Categorical stopping.} Two paradigms are generally used to signal completion. The first is \emph{categorical stopping}, which defines EOS as one token of the model's vocabulary \(\mathcal{V}\) and compares its score directly with the scores assigned to all other \(v \in \mathcal{V}\). The second encodes the EOS signal in a separate binary output \enquote{head} that outputs 1 when the model signals completion. We consider categorical stopping throughout, where the stopping time depends on how the recurrent state reaches the part of the output space that favors EOS.

Under categorical stopping, the comparison between EOS and each content token is linear in the hidden state. Let \(w_v^\top\) denote row \(v\) of \(W_y\). For each content token \(j\), the difference \(C_j = w_{\mathrm{EOS}} - w_j\) defines the EOS comparison
\begin{equation*}
  g_j(h) = C_j^\top h,
\end{equation*}
and stacking the rows \(C_j^\top\) gives \(g(h) = C h\) with \(m(h) = \min_j g_j(h)\), so that \(m_t = m(h_t)\). We call the comparison that attains this minimum the \emph{limiting} EOS comparison. Note that the EOS token is assigned the largest value in the network's output layer exactly when \(m(h) > 0\). Generally, many distinct hidden states lead to the model outputting EOS~\citep{grivas2022unargmaxable,newman2020eos}. Thus, we can define the set \(\mathcal{H}_{\mathrm{EOS}} = \{h \mid m(h) > 0\}\) containing all valid hidden states associated with emission of EOS.

Stopping can then be viewed as a hitting-time problem. Once the model has finished reading all input symbols, its input is set to zero and the hidden-state recurrence becomes autonomous, \(h_{t + 1} = F(h_t)\), and stopping must therefore be encoded by \(h_t\) alone. Hence, chained application of \(F\) determines when the trajectory first enters \(\mathcal{H}_{\mathrm{EOS}}\). We call this first entry time the \emph{EOS hitting time},
\begin{equation*}
  \tau(h) = \min \{u \ge 0 \mid m(F^u(h)) > 0\},
\end{equation*}
so that \(\tau(h)\) counts the steps that remain before EOS. Correct stopping requires \(\tau\) to equal the number of output steps the count instruction specifies, and \(\Delta_t^{J K}\) carries the difference between the two required hitting times along the trajectory.

\paragraph{Metrics.} All accuracies below are measured on held-out sequences. \emph{Stopping accuracy} is the fraction of sequences that emit EOS at the intended step, \emph{content accuracy} is the fraction of content tokens that match the target, and \emph{complete-sequence accuracy} is the fraction of sequences whose entire output, content and EOS position together, matches the target. Where an intervention is applied part way through a rollout, we additionally report \emph{continuation accuracy}, the fraction of reconstructed rollouts whose decoded output matches the one the unmodified network produces.

\section{How the hidden state encodes stopping}%
\label{sec:dynamics}

Fix a model \(\mathcal{M}\), let \(t_c\) denote the step at which it reads the count instruction, and consider two hidden-state rollouts of the same input sequence under count instructions \(J\) and \(K\). The hidden states are identical until the count instruction creates the preactivation difference
\begin{equation*}
  \Delta a_{t_c}^{J K} = W_x(c_J - c_K).
\end{equation*}
Because subsequent inputs are zero, the autonomous recurrence needs to carry this difference until EOS is output.

\subsection{RNN}

During autonomous output, both RNN trajectories satisfy
\begin{equation*}
  a_{t + 1}^{\kappa} = W_h h_t^{\kappa},
  \qquad
  h_{t + 1}^{\kappa} = \tanh(a_{t + 1}^{\kappa}),
  \qquad
  \kappa \in \{J,K\}.
\end{equation*}
For scalar preactivations \(a\) and \(b\), we define the tanh secant
\begin{equation*}
  s(a,b) =
  \begin{cases}
    \dfrac{\tanh(a) - \tanh(b)}{a - b}, & a \neq b, \\[4pt]
    1 - \tanh^2(a),                     & a = b.
  \end{cases}
\end{equation*}
Now let \(D_{t + 1}^{J K}\) be the diagonal matrix containing these secants for the corresponding preactivations in the two rollouts, and define the resulting change in the EOS comparisons as \(\Delta g_t^{J K} = g(h_t^J) - g(h_t^K) = C\Delta_t^{J K}\). This difference then follows the exact recurrence
\begin{align}
  \Delta_{t + 1}^{J K} &= D_{t + 1}^{J K} W_h \Delta_t^{J K},
  \label{eq:rnn-response} \\
  \Delta g_{t + 1}^{J K} &= C D_{t + 1}^{J K} W_h \Delta_t^{J K}. \nonumber
\end{align}
This identity holds for any pair of trajectories.
From its injection through \(W_x(c_J - c_K)\), this difference moves along the
autonomous hidden trajectory before affecting the EOS comparisons
(\cref{fig:rnn-cue-response}). Across the five RNNs, the first two components of an uncentered PCA basis obtained from separate sequences contain on average \(74.5\%\) of the energy of \(\Delta_t^{J K}\). Projected onto these components, the differences on new sequences trace coherent curves, with the two adjacent count changes following nearly the same path (\cref{fig:opening-mechanisms}\textbf{A}).

\begin{figure}[t]
  \centering
  \includegraphics[width=\textwidth]{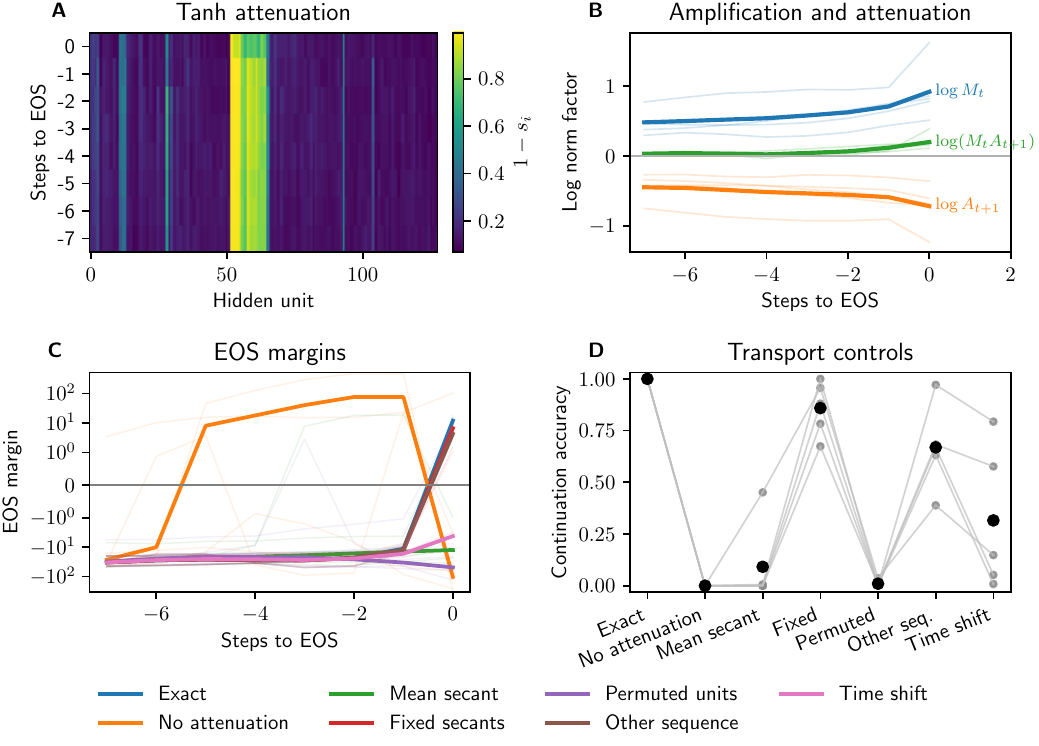}
  \caption{Transport of the RNN trajectory difference. \textbf{A}, mean tanh secants
  for model seed 5 and count change \(4 \to 2\), with hidden units ordered
  by the step of strongest attenuation on separate sequences. \textbf{B},
  multiplication by \(W_h\) expands the trajectory difference before tanh attenuates it.
  Faint lines denote model means and solid lines denote the cohort mean.
  \textbf{C}, EOS margins along the two trajectories. \textbf{D}, continuation
  accuracy under the exact secants and their controlled replacements.}
  \label{fig:rnn-transport}
\end{figure}

The transport of \(\Delta_t^{J K}\) consists of a fixed mixing step followed by a state-dependent scaling step. At each step, \(W_h\) first mixes the current difference. Then, \(D_{t + 1}^{J K}\) scales each hidden unit by a value in \((0, 1]\). Since \(s(a, b)\) depends on the two current states, this scaling changes with the stored sequence, the repeat count, and the output position. To test which of these dependencies matter, we replace the secants in \cref{eq:rnn-response} and propagate \(\Delta_t^{J K}\) with the replaced secants. Note that these replacements run open loop: the propagated difference is decoded through \(W_y\), but the network itself never receives the resulting states. Removing the scaling altogether, or moving each secant to a different hidden unit, leaves almost no correct continuation (\cref{tab:rnn-controls,fig:rnn-transport}). Shifting the same secants in time retains a higher continuation accuracy. Secants taken from other sequences retain more, and their fitting-set mean at the same step yields the highest continuation accuracy of all. Hidden-unit identity and temporal order thus have the largest effects, while the values themselves are largely shared across sequences. The fixed mixing by \(W_h\) admits a further separation into normal and nonnormal parts, neither of which reproduces the stopping times on its own (\cref{app:experimental-details}).

\begin{table}[t]
  \centering
  \small
  \caption{Replacements for the tanh secants \(D_t^{J K}\) in
  \cref{eq:rnn-response}, evaluated across the five RNNs and the three count
  changes. Each replacement propagates \(\Delta_t^{J K}\) over the final copy
  of the shorter target. We add the result to the trajectory of the longer
  count and decode it through \(W_y\), without rerunning the network.
  Continuation accuracy is the fraction of these reconstructions whose output
  matches the output of the unmodified network. The exact row is a numerical check.}
  \label{tab:rnn-controls}
  \begin{tabular}{lc}
    \toprule
    Secants used in the transport & Continuation accuracy \\
    \midrule
    Exact                                    & \(1.000\) \\
    Identity, \(D_t^{J K} = I\)              & \(0.000\) \\
    Permuted across hidden units             & \(0.010\) \\
    Mean across hidden units, same sequence and step & \(0.091\) \\
    Circularly shifted in time               & \(0.316\) \\
    Taken from other sequences               & \(0.669\) \\
    Fitting-set mean, same step              & \(0.859\) \\
    \bottomrule
  \end{tabular}
\end{table}

The transported difference separates trajectories by the number of output steps that remain, which we describe by the current position within the copied sequence and the number of copies remaining. Indexing the output states from \(t = 0\), for a count instruction \(K\) let
\begin{equation*}
  H_{n, p}^{K}(S) = h_{(K - n)L + p}^{K}(S),
\end{equation*}
with \(p \in \{0, \dots, L - 1\}\) the index within the current copy of \(S\) and \(n\) copies of \(S\) remaining. Applying the network's autonomous transition function to \(H_{n, p}^{K}(S)\),
\begin{equation}
  F\!\left(H_{n, p}^{K}(S)\right)
  =
  \begin{cases}
    H_{n, p + 1}^{K}(S), & p < L - 1, \\
    H_{n - 1, 0}^{K}(S), & p = L - 1.
  \end{cases}
  \label{eq:rnn-bilevel-map}
\end{equation}
\Cref{eq:rnn-bilevel-map} describes the change in \(n\) and \(p\) within a single rollout. The position advances until the copy ends, and the number of copies remaining then decreases by one. Because \(L\) is fixed, \((n, p)\) is equivalent to the number of output steps remaining, which equals the hitting time \(\tau(H_{n, p}^{K}(S)) = nL - p\) of \cref{sec:setup}.

\begin{figure}[t]
  \centering
  \includegraphics[width=\textwidth]{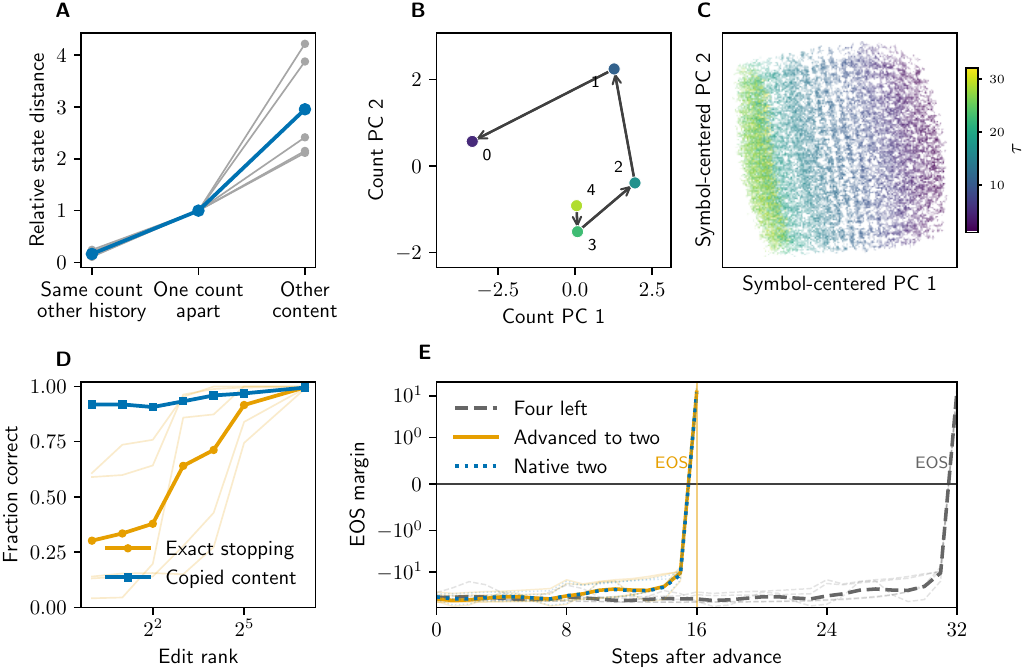}
  \caption{RNN states indexed by the number of output steps remaining. \textbf{A}, distances between states that share \((n, p)\)
  and \(S\) under different count instructions, between states one copy apart,
  and between states storing other content, each relative to the distance
  between states one copy apart. Gray lines denote models and the blue line
  denotes their mean. \textbf{B}, cycle centers projected onto two principal
  components obtained from separate sequences. Arrows follow one complete copy.
  \textbf{C}, evaluation states of model seed 3, the model with the median
  \(R^2\), after subtracting the fitting-set mean state of each emitted symbol,
  projected onto the first two principal components of the fitting states and
  colored by \(\tau\).
  \textbf{D}, stopping and content accuracy after the affine state transformation of
  \cref{eq:remaining-edit} as a function of its rank. \textbf{E}, EOS margins
  after the native recurrence advances a state by two copies.}
  \label{fig:rnn-remaining-count}
\end{figure}

Rollouts that began with different count instructions can contain the same value for \(n\). A model that is asked for four copies of the input, for example, reaches \(n = 2\) after completing two copies. The same model asked for three copies reaches \(n = 2\) after completing one copy of \(S\). States that agree on \((n, p)\) and \(S\) should then coincide regardless of the count instruction:
\begin{equation}
  H_{n, p}^{K}(S) \approx H_{n, p}^{K'}(S), \qquad K, K' \ge n.
  \label{eq:history-collapse}
\end{equation}
We measure this proximity across 5 RNN seeds using 512 held-out sequences. For each model, the distance between the states in \cref{eq:history-collapse} is averaged and divided by the mean distance between states with adjacent values of \(n\) (\cref{app:experimental-details}). In all five models, states that agree on \((n, p)\) and \(S\) lie close to each other regardless of the count instruction that produced them. Their mean distance is \(0.165\) of this reference, with full-state cosine between \(0.996\) and \(0.998\) (Panel~\textbf{A} of \cref{fig:rnn-remaining-count}). The largest ratio of this distance to the distance between adjacent values of \(n\) at any count and position is \(0.862\).

If we average the hidden states over the \(L\) positions of each copy, we obtain one cycle center per copy. Cycle centers with the same \(n\) lie even closer together, at \(0.109\) of the distance between cycle centers with adjacent \(n\). Thus, the states agree more closely once the variation within a copy is removed. Panel~\textbf{B} of \cref{fig:rnn-remaining-count} shows the cycle centers ordered by \(n\) in a PCA projection obtained from separate sequences. The individual states carry the number of output steps remaining in a similarly low-dimensional form once the mean state of each emitted symbol, which dominates their leading principal components, is subtracted. A linear readout from the first two principal components of these centered states then recovers \(\tau\) on new sequences with \(R^2\) between \(0.73\) and \(0.98\) across the five models, which exceeds every one of 32 random two-dimensional projections per model (Panel~\textbf{C} of \cref{fig:rnn-remaining-count}).

Subtracting the cycle center from each state leaves the motion through \(p\). We call the resulting sequence of centered hidden states across the \(L\) positions of \(S\) a \emph{centered path}, since it describes how the state moves within one copy. The position-wise average of the centered paths from the other copies of the same held-out sequence recovers \(0.872\) of the variation of each path. Centered paths of successive copies of the same sequence have a mean cosine of \(0.939\). Shifting the path by one position drops this cosine to \(-0.138\), and a path from another sequence gives \(-0.003\). Thus, the centered path follows \(p\) and depends on \(S\). This is consistent with a position that advances through \(S\) and resets after each complete copy, while the number of copies remaining decreases. Note that it is equally consistent with a single countdown of \(\tau\).

Moving a state from \(n\) to \(n - 1\) should remove one copy from the future
output. We test this with an affine state transformation learned from separate fitting
sequences. For each sequence \(S_i\) in the fitting set, we take the source
state \(h_i\) and the target state \(h_i^-\) at the same output step under
adjacent count instructions:
\begin{equation*}
  h_i = H_{n,0}^{K}(S_i),
  \qquad
  h_i^- = H_{n - 1,0}^{K - 1}(S_i),
  \qquad
  \delta_i = h_i^- - h_i.
\end{equation*}
Let \(\bar h\) and \(\bar\delta\) denote the corresponding means. A rank \(k\)
linear map \(M_k\) then estimates the centered difference from the centered
source state, which yields the state transformation
\begin{equation}
  T_k(h) = h + \widehat{\delta}_k(h), \qquad \widehat{\delta}_k(h) = \bar\delta + M_k(h - \bar h).
  \label{eq:remaining-edit}
\end{equation}
We evaluate \(T_k\) on new sequences by giving it only the hidden
state \(h\) that lies at a copy boundary, i.e., after one copy has completed
and another is started at the next time step. The unchanged RNN outputs EOS one
copy earlier when continuing from \(T_{128}(h)\). The tokens that the RNN
copies before EOS are not affected by the transformation
(\cref{tab:interventions,fig:rnn-remaining-count}). \(T_{128}\) does not
isolate the number of copies remaining as a separate variable, since it acts
on the complete hidden state. The complete target is reproduced less often
when we only add the mean change of the fitting sequences with the same count
and cycle (\cref{tab:interventions}). Thus, most of the change that removes one
copy is shared by all sequences. We measure how much \(T_{128}\) depends
on the stored sequence by using two further controls. The first control
shuffles the target states across the fitting sequences, and the second
control replaces \(h\) by the mean target state. Neither control produces a
single complete output. The accuracy of the copied content then drops to
roughly the level expected by chance over the sixteen content tokens. Both
controls keep the \(-h_i\) term in \(\delta_i = h_i^- - h_i\). Hence, they only
remove the relation between the stored sequence and its target state.

\begin{table}[t]
  \centering
  \small
  \caption{Interventions on the stopping mechanism, evaluated on held-out
  sequences. All accuracies are computed over every rollout. RNN entries are
  means over the five models, and SSM entries are ranges across the five
  models. The fixed-direction rows pool the targets \(r = 1, 2, 4, 5\),
  starting from a rollout of three copies. Since the SSM rotations preserve
  almost all of the copied content, their complete-sequence accuracy rounds to
  the same range as their stopping accuracy.}
  \label{tab:interventions}
  \begin{tabular}{lccc}
    \toprule
    Intervention & Stopping & Content & Complete \\
    \midrule
    \multicolumn{4}{l}{\emph{RNN, affine state transformation \(T_{128}\) from \(n\) to \(n - 1\) copies remaining}} \\
    Fitted transformation, new sequences & \(0.995\) & \(0.994\) & \(0.944\) \\
    Mean change, same count and cycle   & \(0.829\) & \(0.976\) & \(0.728\) \\
    Targets shuffled across sequences   & \(0.456\) & \(0.049\) & \(0.000\) \\
    \(h\) replaced by mean target state & \(0.520\) & \(0.046\) & \(0.000\) \\
    \midrule
    \multicolumn{4}{l}{\emph{SSM, amplitude-preserving phase rotations}} \\
    Independent rotation per plane & \(0.015\)--\(0.047\) & \(0.999\)--\(1.000\) & \(0.015\)--\(0.047\) \\
    Rotation shared by all planes  & \(0.268\)--\(0.316\) & \(1.000\)                & \(0.268\)--\(0.316\) \\
    \midrule
    \multicolumn{4}{l}{\emph{SSM, fixed direction \((r - 3)\xi\) at the first output state, \(r \in \{1, 2, 4, 5\}\)}} \\
    Fitted direction \(\xi\)               & \(0.911\)--\(0.976\) & \(0.998\)--\(1.000\) & \(0.911\)--\(0.974\) \\
    Mean count-instruction difference      & \(0.548\)--\(0.724\) & \(0.950\)--\(0.991\) & \(0.547\)--\(0.724\) \\
    Norm-matched random direction          & \(0.000\)--\(0.001\) & \(0.953\)--\(0.975\) & \(0.000\)--\(0.001\) \\
    \bottomrule
  \end{tabular}
\end{table}

\FloatBarrier
\subsection{Linear SSM}

\begin{figure}[t]
  \centering
  \includegraphics[width=0.5\textwidth]{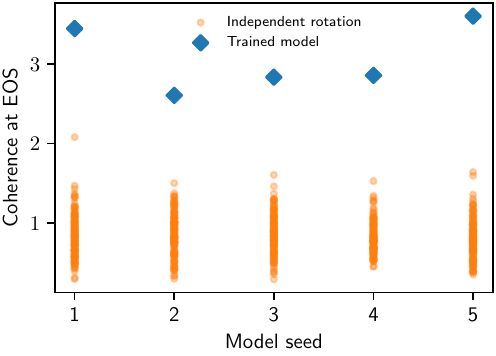}
  \caption{Coherence of the rotation-plane contributions at EOS in each SSM,
  compared with independent amplitude-preserving rotations of the planes.
  Stopping and content accuracy under these rotations are given in
  \cref{tab:interventions}.}
  \label{fig:ssm-phase}
\end{figure}

The linear SSM defined in \cref{eq:ssm} carries the trajectory difference through a fixed linear recurrence. After \(u\) autonomous output steps,
\begin{equation*}
  \Delta_{t_c + u}^{J K} = A^u W_x(c_J - c_K),
  \qquad
  \Delta g_{t_c + u}^{J K} = C A^u W_x(c_J - c_K).
\end{equation*}
Each transition block has the form \(A_q = \rho_q\Rot(\omega_q)\). Let \(P_q\)
select block \(q\), and let \(r_q^{J K}\) and \(\phi_q^{J K}\) be the amplitude
and phase of \(P_qW_x(c_J - c_K)\). The trajectory difference in this plane is
\begin{equation*}
  P_q\Delta_{t_c + u}^{J K}
  =
  \rho_q^u r_q^{J K}
  \begin{bmatrix}
    \cos(\phi_q^{J K} + u\omega_q) \\
    \sin(\phi_q^{J K} + u\omega_q)
  \end{bmatrix}.
\end{equation*}
The output matrix sums the contribution \(v_{q,u}^{J K}\) from every plane:
\begin{equation*}
  \Delta g_{t_c + u}^{J K}
  =
  \sum_{q = 1}^{64}
  \underbrace{\rho_q^u r_q^{J K} C P_q^\top
  \begin{bmatrix}
    \cos(\phi_q^{J K} + u\omega_q) \\
    \sin(\phi_q^{J K} + u\omega_q)
  \end{bmatrix}}_{v_{q,u}^{J K}}.
\end{equation*}
The fixed relative phases make the plane contributions cancel during copying
and align near the stopping time, increasing the effect of the count instruction
on the EOS comparisons. Their coherence
\(\lVert \sum_q v_{q,u}^{J K}\rVert_2 /
(\sum_q \lVert v_{q,u}^{J K}\rVert_2^2)^{1/2}\) lies between
\(2.61\) and \(3.60\) at EOS across the five models (\cref{fig:ssm-phase}).

Amplitude and phase are separable, meaning we can intervene on them independently. Rotating the trajectory difference within each plane on its own (\cref{app:experimental-details}) preserves
every plane amplitude and changes only the phases relative to the other planes,
which removes almost all of the intended stopping behavior while leaving the
copied content intact (\cref{tab:interventions}). In contrast, a rotation shared by all
planes keeps the phases of the planes relative to each other, i.e., every plane
is rotated by the same angle. The model then still stops at the intended step
in between \(0.268\) and \(0.316\) of the rollouts (\cref{tab:interventions}). Thus, the
relative phase across planes is necessary for correct EOS timing. Note that it
is not sufficient, since the shared rotation keeps every relative phase and
still removes most of the intended stopping. Hence, the orientation of the
trajectory difference relative to \(W_y\) matters as well. Both interventions keep the
amplitude of every plane fixed. Thus, they cannot be used to test whether the
amplitudes contribute to EOS timing.

Because the SSM is linear, the stopping time can also be set directly by adding
a fixed direction to the hidden state. For each model, we fit one direction
\(\xi\) such that adding \((r - 3)\xi\) to the first output state of a rollout with
three copies produces \(r\) copies of \(S\), followed by EOS. We fit \(\xi\) on separate sequences for \(r \in \{1, 2, 5\}\) by requiring the target token to win every future comparison by a margin of \(0.1\), and hold out \(r = 4\) (\cref{app:experimental-details}). On new sequences, the same direction sets the number of copies for every \(r\) from 1 to 5, including \(r = 5\), one copy more than the model was trained on (\cref{tab:interventions}). The held-out \(r = 4\) is reached in every sequence of every model. The direction meets the pre-specified success criteria, a complete-sequence accuracy of at least \(0.75\) at every target and \(0.9\) on average, in four of the five models, and it fails at \(r = 6\). Moving the state along the mean difference that the count instruction itself produces per copy reaches the target less often, and norm-matched random directions almost never do.

\FloatBarrier
\subsection{What the two mechanisms share}

Because both architectures use a linear decoder \(W_y\) to generate the EOS token, each must associate the count with EOS emission. We find that the models
achieve this association via opposite changes in the trajectory difference. We measure its magnitude as \(\lVert \Delta_t^{J K} \rVert_2\) and its visibility through the EOS comparisons as \(\lVert C \Delta_t^{J K} \rVert_2 / \lVert \Delta_t^{J K} \rVert_2\).
During the last copy of \(S\), the trajectory difference grows in the RNN by factors between \(1.52\) and \(2.16\) and contracts in the SSM by factors between \(0.61\) and \(0.65\). Nonetheless, visibility rises in four of the five RNNs, by factors between \(1.08\)
and \(5.52\), and in every SSM, by factors between \(1.12\) and \(1.29\). Every model also reaches EOS with positive alignment to the limiting comparison.
The cosines lie between \(0.214\) and \(0.622\) in the RNN and between
\(0.527\) and \(0.580\) in the SSM (\cref{fig:guard-exposure}). The trajectory difference thus has to
become visible to \(C\) at the intended step, while its magnitude at that step
remains a property of the recurrent transition. A separate comparison shows that choosing among several stopping times requires more principal components than copying with a fixed stopping time (\cref{app:task-compression,fig:short-reduction}).

\FloatBarrier
\section{Discussion}%
\label{sec:discussion}

Stopping with an EOS token requires the nonlinear RNNs and linear SSMs studied here to encode a timing signal in their hidden states. Using the repeat-copy task, this signal is visible as a difference between trajectories generated from the same sequence under different repeat counts. The difference persists throughout the model's autonomously generated output, including steps at which the EOS margin remains negative. Its presence at these steps shows that the hidden dynamics contain information about the future stopping time before this information can affect the current output. The EOS margin \(m_t\) thus acts as an observable of the effect of the model's hidden dynamics, which implement the timing computation.

The form of this timing computation depends on the recurrent transition. The RNNs we consider here transport the trajectory difference through repeated mixing by their recurrent matrix \(W_h\) and state-dependent scaling by the activation function's secants. Their hidden states carry the remaining step count \(\tau = nL - p\), which at the fixed sequence length is equivalent to the pair of \(n\), the number of copies of \(S\) remaining, and \(p\), the current position within \(S\), so that our experiments do not determine whether the network represents the two indices separately. An affine state transformation estimated on separate sequences shifts a state at a copy boundary to the states with one fewer copy remaining, and the unmodified network then ends its autonomous output one copy earlier while copying every preceding token correctly.

We find that the linear SSM transports \(\Delta_t^{J K}\) through fixed rotation planes. Contributions from these planes cancel in the EOS comparisons during copying before reinforcing near the stopping time. Independent phase rotations preserve the response amplitudes and copied symbols; most of the intended stopping behavior disappears. A shared rotation retains more stopping behavior because it preserves the relative phases. Relative phase across planes and the orientation of the response toward \(W_y\) therefore both govern when the trajectory difference affects the EOS comparisons. A single fixed direction in the SSM state also sets the stopping time, to any number of copies from one to five.

\section*{Limitations}

Because \(L\) is fixed, there exists a bijection from \((n, p)\) to the remaining step count \(\tau = n L - p\), making our nested description of the RNN's dynamics indistinguishable from a single counter of the remaining number of output steps. The affine state transformation, applied at copy boundaries, does not distinguish these either. Varying sequence length independently of repeat count would distinguish the aforementioned axes. Besides, the explicit count tokens and deterministic sequences also make the timing response identifiable. The analysis in this paper concerns instructed halting in small tanh RNNs and linear SSMs. Experiments using gated RNNs and nonlinear SSMs will be needed to determine whether the same forms of transport occur in other recurrent architectures.

\bibliography{references}
\bibliographystyle{plainnat}

\appendix

\section{Experimental details}%
\label{app:experimental-details}

\begin{figure}[t]
  \centering
  \includegraphics[width=0.86\textwidth]{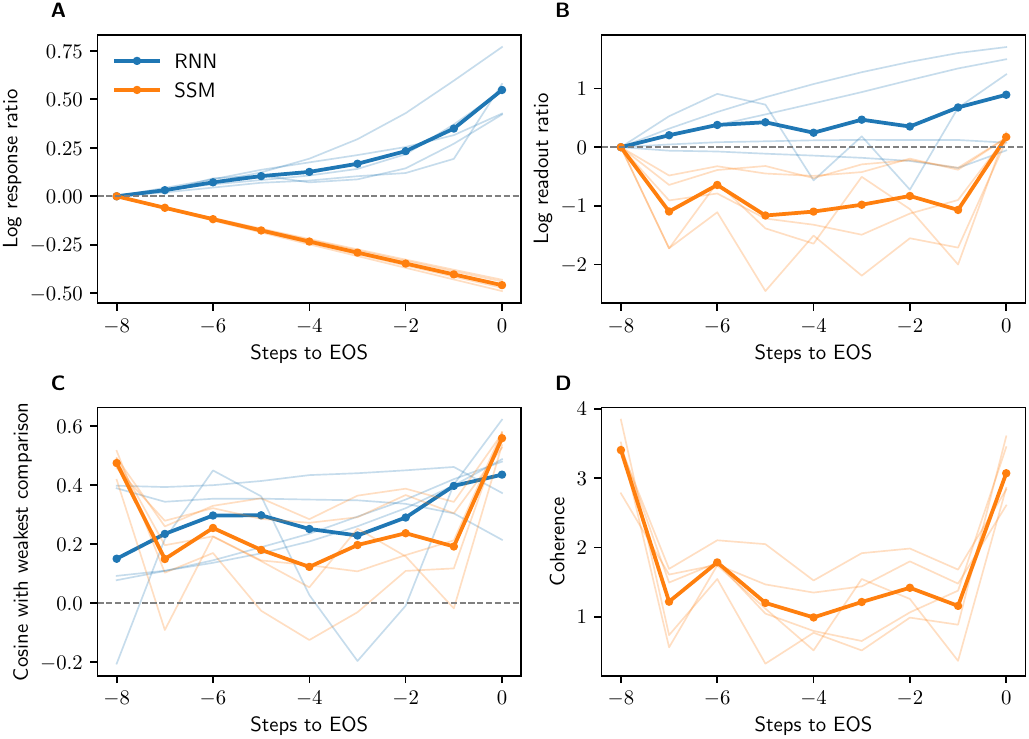}
  \caption{Magnitude of \(\Delta_t^{J K}\) and its visibility through the EOS
  comparisons. \textbf{A}, magnitude of \(\Delta_t^{J K}\) relative to the start of the
  last copy. \textbf{B}, visibility through all EOS comparisons.
  \textbf{C}, cosine with the weakest EOS comparison at the target state.
  \textbf{D}, coherence of the SSM plane contributions. Faint lines denote
  models and dark lines denote their means.}
  \label{fig:guard-exposure}
\end{figure}

The nonlinear RNNs we use to train on repeat-copy have width 128 and use the default dense PyTorch initialization for \(W_h\). The RNNs are optimized for 100,000 updates and the SSMs for 5,000, chosen based on how long they took to converge. We train both architectures with Adam~\citep{kingma2014adam} using batch size 128, unit gradient clipping, and a cosine learning rate schedule from \(10^{-3}\) to \(10^{-5}\) over 100,000 updates. Training uses the cross-entropy loss. We do not provide the models with a countdown signal or output feedback. All training, fitting, evaluation, and control streams are derived from the single data seed 123, and the five models of each cohort use the model seeds 1 to 5.

\paragraph{RNN transport of the trajectory difference.} We evaluate five RNN seeds, 1 through 5, for the count changes \(3 \to 2\), \(4 \to 2\), and \(4 \to 3\) from the shorter target's final copy through EOS. Separate streams supply 256 fitting sequences, 512 evaluation sequences, and the hidden unit permutations, content exchanges, and temporal shifts. The cycle comparison uses the same fitting and evaluation sequences. It reuses eight position-specific maps from the first copy at later copies and includes an absolute time control fitted separately at every output step.

\begin{figure}[H]
  \centering
  \includegraphics[width=\textwidth]{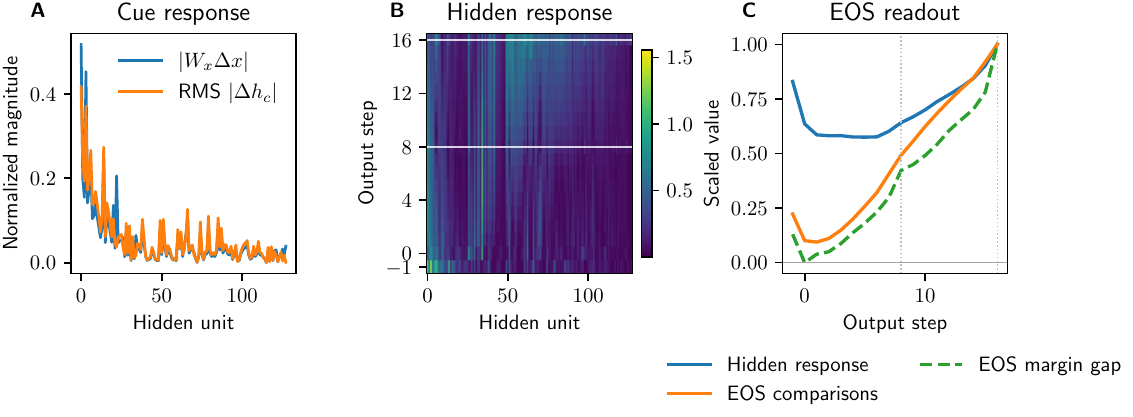}
  \caption{Propagation of the RNN trajectory difference from the instruction to the
  EOS comparisons for model seed 5 and count change \(4 \to 2\).
  \textbf{A}, the preactivation response \(W_x(c_J - c_K)\) and resulting
  hidden response at the instruction. Hidden units are ordered by the step of
  their largest response on separate sequences. \textbf{B}, the hidden
  response during autonomous output, with horizontal lines at copy boundaries.
  \textbf{C}, the norm of \(\Delta_t^{J K}\), its effect on the EOS comparisons,
  and the difference between the two EOS margins, each scaled by its largest
  absolute value.}
  \label{fig:rnn-cue-response}
\end{figure}

\paragraph{SSM phase interventions.} The intervention parameters are fixed in advance, and we evaluate seeds 1 through 5 on new sequences and phase draws. The phase controls
rotate \(\Delta_t^{J K}\) at 8, 4, or 1 steps before the target EOS, using either an independent rotation for each plane or one
common rotation for all planes, with angles drawn uniformly from \((-\pi, \pi)\). Neither control uses output weights or
token labels. Stopping accuracy is computed over all rollouts, without filtering. The fixed direction \(\xi\) is fitted on 64 sequences and evaluated on 256 new sequences, against 32 norm-matched random directions per model.

\paragraph{Remaining copy geometry.} RNN seeds 1 through 5 use the same 256 fitting and 512
evaluation sequences. A wider cohort, again with model seeds 1 through 5, is
trained on repeat counts from one to six and evaluated on the same sequences. We estimate the affine state transformation from the fitting sequences and apply it at copy boundaries in the evaluation sequences. The cycle centers and centered paths use the same 512
evaluation sequences, with the principal component axes obtained from the 256
fitting sequences. For each sequence, the history-collapse statistic averages
distances between states that share \((n, p)\) under different initial counts and
normalizes them by the distance between adjacent values of \(n\). For the remaining-step readout, the symbol means, the principal components, and the linear readout of \(\tau\) are all estimated on the fitting sequences and evaluated on the evaluation sequences, and the random projections are 32 uniformly drawn orthonormal two-dimensional bases per model. We predict each centered path from the other copies of the same sequence. The controls shift its position by one or substitute a different sequence. Of the wider cohort, four models solve every count from one to six. The fifth, seed 2, does not and is excluded. Two of the four, seeds 1 and 3, show the same state organization and the same causal edit at every count, while the other two keep substantial history dependence at high counts. The main cohort does not exclude any model. The stronger state-transformation
controls use the same sequences and a separate stream for the target permutations.

\paragraph{Task compression.} The cross-width comparison uses model seeds 1 through 5. The short-copy cohort uses model seeds 6 through 10, since seeds 1 through 5 would reproduce the width-four models of the cross-width comparison.
For short copy, the principal component basis contains only \((-1, -1)\) and \((+1, +1)\).
The reduced rollout is evaluated on the held-out inputs \((-1, +1)\) and \((+1, -1)\), against sixty-four seeded Haar control bases per model. All four inputs enter the native competence check and the state factorization.

\paragraph{Transport verification.}

Using \(t_c\) for the count input step, at step \(t\), \(\Delta_t^{J K}\) changes the EOS comparisons by
\begin{equation*}
  \Delta g_t^{J K}
  = C D_t^{J K} W_h D_{t - 1}^{J K} W_h \cdots
  D_{t_c}^{J K} W_x(c_J - c_K).
\end{equation*}
Across the float32 trajectories in \cref{fig:rnn-transport}, the maximum state
and EOS comparison discrepancies are \(9.40 \times 10^{-6}\) and
\(8.43 \times 10^{-5}\). The discrepancies measure the numerical agreement
between the direct network rollout and the exact pairwise factorization.

\paragraph{Normal and nonnormal mixing.}
The transport in \cref{eq:rnn-response} combines fixed mixing by \(W_h\) with
state-dependent scaling by \(D_{t + 1}^{J K}\). To examine the fixed mixing, we obtain a spectrum-preserving normal surrogate \(W_N\) from the real Schur decomposition by replacing each two-dimensional block with a scaled rotation that has the same conjugate eigenvalues. Its residual \(W_{\mathrm{res}} = W_h - W_N\) separates the response as
\begin{equation*}
  \Delta_{t + 1}^{J K}
  = D_{t + 1}^{J K} W_N \Delta_t^{J K} + D_{t + 1}^{J K} W_{\mathrm{res}} \Delta_t^{J K},
  \qquad
  W_N^\top W_N = W_N W_N^\top.
\end{equation*}
Across the five RNNs and four fixed Schur orderings, \(W_N\) and \(W_{\mathrm{res}}\) have
mean Frobenius norms of \(0.357\) and \(0.931\) relative to \(W_h\). After tanh
scaling, their contributions have root mean square norms of \(0.863\) and
\(0.391\) relative to the complete next response. The normal surrogate alone
preserves \(0.019\) of the required EOS hitting times. The residual alone preserves \(0.0004\). Their sum recovers every hitting time. Removing either
component from the autonomous recurrence reduces complete-sequence accuracy to
zero in every model. The normal surrogate carries most of the one-step count
response. The residual supplies the correction required for the hidden
difference to reach the EOS region.

\paragraph{Advancing a trajectory.}

For hidden state \(h = H_{n,p}^{K}(S)\) on an unmodified network trajectory and a
desired remaining count \(1 \le n^\star \le n\), let
\(\ell = (n - n^\star)L\). Advancing the autonomous recurrence by \(\ell\) steps reaches \(H_{n^\star,p}^{K}(S)\), as shown in the following:
\begin{equation}
\begin{aligned}
  \Phi_{n \rightarrow n^\star}\!\left(H_{n,p}^{K}(S)\right) := F^\ell\!\left(H_{n,p}^{K}(S)\right) &= h_{(K - n)L + p + (n - n^\star)L}^{K}(S) \\
          &= H_{n^\star,p}^{K}(S).
\end{aligned}
\label{eq:remaining-cycle}
\end{equation}
Under \cref{eq:remaining-cycle}, the same trajectory advances by complete
copies without changing the stored sequence or current position. The remaining
stopping time then changes from \(nL - p\) to \(n^\star L - p\):
\begin{equation*}
  \tau\!\left(H_{n,p}^{K}(S)\right) = n L - p,
  \qquad
  \tau\!\left(\Phi_{n \rightarrow n^\star}(H_{n,p}^{K}(S))\right)
  = n^\star L - p.
\end{equation*}
At any boundary between copies of \(S\), the endpoint map \(\Phi_{n \rightarrow 0} = F^{nL}\) reaches
the EOS state through the unmodified recurrence. The affine state transformation in
\cref{eq:remaining-edit} approximates the one-copy transition
\(\Phi_{n \rightarrow n - 1}\) from separate sequences and applies the resulting
state relation to new content.

The SSM implements this advancement with the fixed matrix \(A^\ell\). For the RNN,
let \(h^{(0)} = h\), \(a^{(j + 1)} = W_h h^{(j)}\), and
\(h^{(j + 1)} = \tanh(a^{(j + 1)})\). Define
\(\alpha(a) = \tanh(a)/a\) for \(a \ne 0\), \(\alpha(0) = 1\), and
\(G(a) = \operatorname{diag}(\alpha(a_1), \ldots, \alpha(a_d))\). Repeated substitution
expresses \(F^\ell(h)\) as a matrix that depends on the current state:
\begin{equation*}
  F^\ell(h)
  = \underbrace{[G(a^{(\ell)})W_h]\cdots[G(a^{(1)})W_h]}_{\mathcal A_\ell(h)}h.
\end{equation*}
The entries of \(\mathcal A_\ell(h)\) depend on the intermediate states along the
chosen trajectory. For \(\ell = 16\), this matrix maps a state with four copies
remaining to the state reached two copies later. Across all five models and 512
evaluation sequences per model, the unchanged recurrence produces fifteen
negative EOS margins followed by a positive margin on step sixteen. Every
intervening content token remains correct except one, which RNN seed 4 also miscopies without the transformation.

The exact cycle transformation defines the target state
\(h^\star = \Phi_{n \rightarrow n^\star}(h)\). For \(h \ne 0\), a rank-one change
to the recurrent matrix sends \(h\) to the successor of this target in one
update:
\begin{equation*}
  \Delta W_h^\star
  = \frac{W_h(h^\star - h)h^\top}{h^\top h}.
\end{equation*}
For this choice,
\(\tanh((W_h + \Delta W_h^\star)h) = \tanh(W_h h^\star) = F(h^\star)\).
After a single edited update reaches \(F(h^\star)\), restoring \(W_h\)
continues along the original dynamics from this successor. The rank-one
construction depends on the current state and target. The affine state transformation in \cref{eq:remaining-edit} is fitted on separate sequences and applies across content.

\section{Reduced dynamics for simpler tasks}%
\label{app:task-compression}

\begin{figure}[H]
  \centering
  \includegraphics[width=\textwidth]{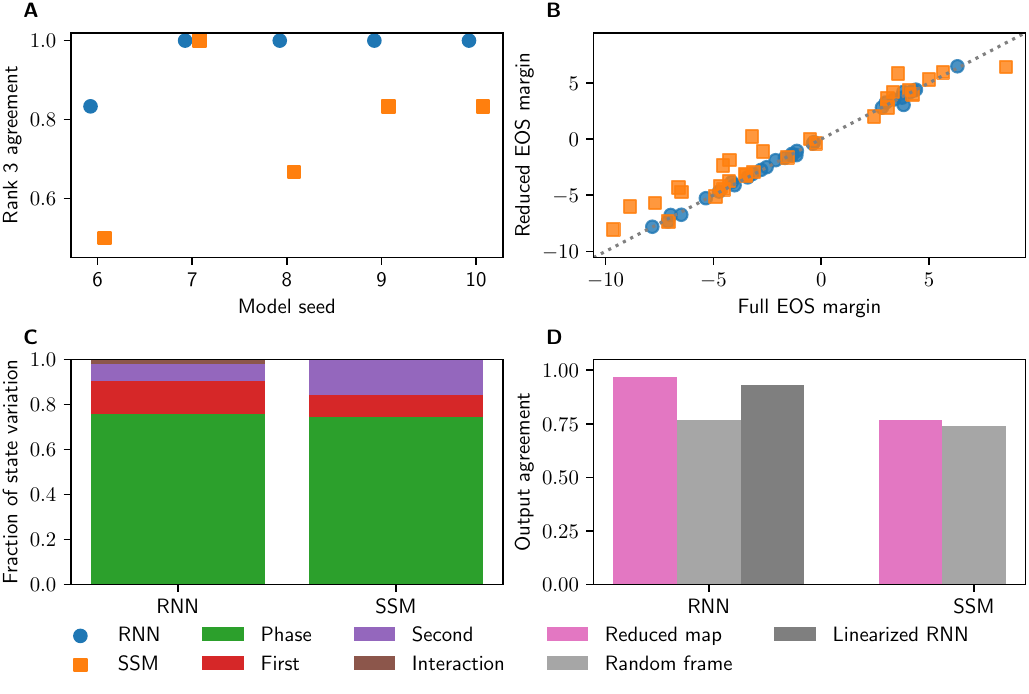}
  \caption{Fixed short-copy dynamics under reduced recurrences. \textbf{A},
  agreement of the rank three system across five models. \textbf{B}, EOS
  margins from the full and reduced systems on the mixed sequences.
  \textbf{C}, contributions of output position and symbols to the exhaustive
  state table. \textbf{D}, output agreement for the reduced system, random
  subspaces, and the linearized RNN. Rank three preserves every output in four
  of the five RNNs and the stopping time in four of the five SSMs.}
  \label{fig:short-reduction}
\end{figure}

We compare fixed short-copy and cued-delay tasks by projecting the first output
state onto a principal component basis obtained from separate trajectories,
then advancing it through the original recurrence. The smallest rank that
reproduces every target token defines the autonomous projection rank. At widths
\(4, 8, 16,\) and \(32\), the median short-copy ranks are \(2, 3, 3,\) and
\(3\) in the RNN and \(3, 4, 5,\) and \(4\) in the SSM. The corresponding cued-delay ranks are \(4, 6, 14,\) and \(17\) in the RNN and \(4, 7, 11,\) and
\(32\) in the SSM. The cued-delay rank exceeds the short-copy rank at every width in both architectures under the exact output criterion
(\cref{fig:short-reduction}).

At width four, both architectures solve the short-copy and cued-delay tasks.
For the fixed three-output copy program, let \(\mu\) be the mean of the six RNN
states produced by the homogeneous sequences \((-1, -1)\) and \((+1, +1)\). The first three
singular directions of these states form \(B \in \mathbb{R}^{4 \times 3}\), with
\(\xi_t = B^\top(h_t - \mu)\). The reduced recurrence is
\begin{equation*}
  \xi_{t + 1} = B^\top\!\left[
    \tanh\!\left(W_h(\mu + B \xi_t)\right) - \mu
  \right],
  \qquad
  \hat{y}_t = \argmax_{v} w_v^\top(\mu + B \xi_t).
\end{equation*}
Starting from the first output state, this rank three recurrence reproduces
every output for the separate mixed sequences \((-1, +1)\) and \((+1, -1)\) in four of the five
RNNs. Mean output agreement is \(0.768\) for random rank three subspaces and
\(0.933\) for a first order linearization of tanh at \(\mu\).

The corresponding SSM reduction replaces \(\tanh(W_h h)\) with \(A h\). At
rank three, it preserves the EOS position in four of the five models and reproduces the
complete mixed sequences in one. The full four-dimensional state reproduces
every output in every model. For output position \(p\) and input symbols
\(a, b \in \{-1, +1\}\), the exhaustive state table has the decomposition
\begin{equation*}
  h_p(a, b) = \mu_p + a u_p + b v_p + a b \eta_p.
\end{equation*}
Here \(\mu_p\) is the mean state at position \(p\), \(u_p\) and \(v_p\) are the two symbol effects, and \(\eta_p\) is their interaction. Output position accounts for median fractions \(0.759\) and \(0.743\) of centered variation in the RNN and SSM. The corresponding interaction fractions are \(0.018\) and numerical zero.

\end{document}